\documentclass{sig-alternate}

\usepackage{makeidx}  
\usepackage{algorithmic}
\usepackage{algorithm}
\usepackage{amsmath}
\usepackage{amssymb}
\usepackage{graphicx}
\usepackage{epsf}
\usepackage{enumerate}
\usepackage{color}
\usepackage{paralist}

\newtheorem{thm}{Theorem}

\newtheorem{ex}[thm]{Example}

\newcommand{\E}{{\operatorname{\mathbf{E}}}}  

\newcommand{\mt}{{\top}}  

\newcommand{\vecb}[1]{{\mathbf{#1}}}
\newcommand{\abs}[1]{\left\lvert#1\right\rvert}  
\newcommand{\Cset}{{\mathbb{C}}}  

\newcommand{\defeq}{{\stackrel{\mathrm{def}}{=}}}

\newcommand{\algref}[1]{Algorithm~\ref{#1}}
\newcommand{\figref}[1]{Figure~\ref{#1}}

\newcommand{\secref}[1]{Section~\ref{#1}}
\newcommand{\thmref}[1]{Theorem~\ref{#1}}
\newcommand{\eqnref}[1]{Equation~(\ref{#1})}
\newcommand{\eqnsref}[1]{Equations~(\ref{#1})}
\newcommand{\Aset}{\mathcal{A}}
\renewcommand{\Cset}{\mathcal{X}}
\newcommand{\Aalg}{\mathsf{A}}
\newcommand{\algfont}[1]{\texttt{#1}}

\newcommand{\planfont}[1]{}
\newcommand{\lihong}[1]{}

\long\def\comment#1{}

\begin{document}

\conferenceinfo{WSDM'11,} {February 9--12, 2011, Hong Kong, China.} 
\CopyrightYear{2011}
\crdata{978-1-4503-0493-1/11/02}
\clubpenalty=10000
\widowpenalty = 10000 

\title{Unbiased Offline Evaluation of Contextual-bandit-based News Article Recommendation Algorithms}
%
%
%
%
%

\numberofauthors{1} 
%
\author{
%
%
\alignauthor Lihong Li \,\,\,\,\,\, Wei Chu \,\,\,\,\,\, John Langford \,\,\,\,\,\, Xuanhui Wang\\
       \affaddr{Yahoo! Labs\\ 701 First Ave, Sunnyvale, CA, USA 94089}\\
       \email{\{lihong,chuwei,jl,xhwang\}@yahoo-inc.com}
}



\maketitle

\begin{abstract}
Contextual bandit algorithms
have become popular for online recommendation systems
such as Digg, Yahoo! Buzz, and news recommendation in general.
\emph{Offline} evaluation of the effectiveness of new algorithms
in these applications is critical for protecting online user
experiences but very challenging due to their ``partial-label'' nature.
Common practice is to create a simulator which simulates the
online environment for the problem at hand and then run an
algorithm against this simulator. However, creating simulator
itself is often difficult and modeling bias is usually unavoidably
introduced. In this paper, we introduce a \emph{replay}
methodology for contextual bandit algorithm evaluation.
Different from simulator-based approaches, our method is
completely data-driven and very easy to adapt to different
applications. More importantly, our method can provide provably
unbiased evaluations. Our empirical results on a large-scale
news article recommendation dataset collected from Yahoo!
Front Page conform well with our theoretical results.
Furthermore, comparisons between our offline replay and online
bucket evaluation of several contextual bandit algorithms show
accuracy and effectiveness of our offline evaluation method.
%
\end{abstract}




\section{Introduction} \label{sec:introduction}

Web-based content recommendation services such as Digg, Yahoo!
Buzz and Yahoo! Today Module (Figure~\ref{fig:fptoday})
leverage user activities such as clicks to identify the most attractive
contents. One inherent challenge is how to score newly
generated contents such as breaking news, especially
when the news first emerges and little data are available.
A personalized service which can tailor contents towards individual users
is more desirable and challenging.

A distinct feature of these applications
is their ``partial-label'' nature: we observe user feedback (click or not)
for an article \emph{only when} this article is displayed.
A key challenge thus arises which is known as the exploration/exploitation
tradeoff: on one hand, we \emph{want to exploit} (i.e., choose articles
of higher quality estimates to promote our business of interest), but
on the other than, we \emph{have to explore} (i.e., choose articles with
lower quality estimates to collect user feedback so as to
improve our article selection strategy in the long run).
The balance between exploration and exploitation may be modeled
as a ``contextual bandit''~\cite{Langford08Epoch}---a subclass of reinforcement learning
problems~\cite{Sutton98Reinforcement}, and is also
present in many other important Web-based applications such as online
ads display and search query suggestion, etc.


An ideal way to evaluate a contextual-bandit algorithm is to
conduct a bucket test, in which we run the algorithm to serve
a fraction of live user traffic in the real recommendation system.
However, not only is this method expensive, requiring substantial
engineering efforts in deploying the method in the real system,
but it may also have negative impacts on user experience.
Furthermore, it is not easy to guarantee replicable comparison
using bucket tests as online metrics vary significantly over time.
\emph{Offline} evaluation of contextual-bandit algorighms thus
becomes valuable when we try to optimize an online recommendation
system.

Although benchmark datasets for supervised learning 
such as the UCI repository~\cite{Blake98Repository}
have proved
valuable for empirical comparison of algorithms, collecting
benchmark data towards reliable offline evaluation has been
difficult in bandit problems. In our application of news
article recommendation on Yahoo! Front Page, for example, each
user visit results in the following information stored in the
log: user information, the displayed news article, and user
feedback (click or not). When using data of this form to
evaluate a bandit algorithm offline, we will not have user
feedback if the algorithm recommends a different news article
than the one stored in the log. In other words, data in
bandit-style applications only contain user feedback for
recommendations that were actually displayed to the user, but
not all candidates. This ``partial-label'' nature raises a
difficulty that is the key difference between evaluation of
bandit algorithms and supervised learning ones.

Common practice for evaluating bandit algorithms 
is to create a simulator and then run the algorithm against it.
With this
approach, we can evaluate any bandit algorithm without having to
run it in a real system. Unfortunately, there are 
two major drawbacks with this approach. First, creating a
simulator can be challenging and time-consuming for practical
problems. Second, evaluation results based on artificial
simulators may not reflect the actual performance since
simulators are only rough approximations of real problems
and unavoidably contains modeling bias.

Our contributions are two-fold. First, we describe and study an
offline evaluation method for bandit algorithms, which
enjoys valuable theoretical guarantees including unbiasedness and
accuracy. Second, we verify the method's effectiveness by 
comparing its evaluation results to online bucket results
using a large volume of data recorded from Yahoo! Front Page.  These
positive results not only encourage wide use of the proposed
method in other Web-baesd applications, but also suggest a
promising solution to create benchmark datasets from real-world
applications for bandit algorithms.

\paragraph{Related Work} 
Unbiased evaluation has been studied before under different settings.
While our unbiased evaluation method is briefly sketched in an earlier paper~\cite{Li10Contextual} and may be interpreted as a special case of the \emph{exploration scavenging} technique~\cite{Langford08Exploration}, we conduct a thorough investigation in this work, including improved theoretical guarantees and positive empirical evidence using online bucket data.

\section{Contextual Bandit Problems} \label{sec:notation}

The multi-armed bandit problem~\cite{Berry85Bandit} is a classic and popular model
for studying the exploration-exploitation tradeoff. 
Despite the simplicity of the model, it has found wide applications in important problems like medical treatment allocation, and recently, in challenging, large-scale problems like Web content optimization~\cite{Agarwal09Explore,Li10Contextual}.
Different from the classic multi-armed bandit problems, we are particularly
concerned with a more interesting setting where for each round contextual information
is available for decision making.

\subsection{Notation}

For the purpose of this paper, we consider the multi-armed bandit problem with contextual information.  Following previous work~\cite{Langford08Epoch}, we call it a \emph{contextual bandit problem}.\footnote{In the
literature, contextual bandits are sometimes called bandits with covariate~\cite{Woodroofe79One}, associative reinforcement learning~\cite{Kaelbling94AssociativeFunctions}, bandits with expert advice~\cite{Auer02Nonstochastic}, bandits with side information~\cite{Wang05Bandit}, and associative bandits~\cite{Strehl06Experience}.}
Formally, we define by $\Aset=\{1,2,\ldots,K\}$ the set of arms, and
a contextual-bandit algorithm $\Aalg$ interacts with the
\emph{world} in discrete trials $t=1,2,3,\ldots$.  In trial $t$:
\begin{enumerate}
\item{The world chooses a feature vector $\vecb{x}_{t}$ known as the
  \emph{context}.  Associated with each arm $a$ is a real-valued
  payoff $r_{t,a}\in[0,1]$ that can be related to the context
  $\vecb{x}_{t}$ in an arbitrary way.  We denote by $\Cset$ the (possibly infinite) set of
  contexts, and $(r_{t,1},\ldots,r_{t,K})$ the payoff vector.  Furthermore,
  we assume $(\vecb{x}_t,r_{t,1},\ldots,r_{t,K})$ is drawn i.i.d. from some
  unknown distribution $D$.} 
%
\item{Based on observed payoffs in previous trials and the current context $\vecb{x}_t$, $\Aalg$ chooses an arm $a_t\in\Aset$, and receives payoff $r_{t,a_t}$.  It is important to emphasize here that \emph{no} feedback information (namely, the payoff $r_{t,a}$) is observed for \emph{unchosen} arms $a \ne a_t$.}
\item{The algorithm then improves its arm-selection strategy with all information it observes, $(\vecb{x}_{t,a_t},a_t,r_{t,a_t})$.} \label{def:bandit-feedback}
\end{enumerate}

In this process, the \emph{total $T$-trial payoff} of $\Aalg$ is defined as
\[
G_\Aalg(T) \defeq \E_D\left[\sum_{t=1}^T r_{t,a_t}\right],
\]
where the expectation $\E_D[\cdot]$ is defined w.r.t. the i.i.d. generation
process of $(\vecb{x}_t,r_{t,1},\ldots,r_{t,K})$ according to distribution $D$
(and the algorithm $\Aalg$ as well if it is not deterministic).
Similarly, given a policy $\pi$ that maps contexts to actions, $\pi:\Cset\mapsto\Aset$, we define its total $T$-trial payoff by
\[
G_\pi(T) \defeq \E_D\left[\sum_{t=1}^T r_{t,\pi(\vecb{x}_t)}\right]=T\cdot\E_D\left[r_{1,\pi(\vecb{x}_1)}\right],
\]
where the second equality is due to our i.i.d. assumption.
Given a reference set $\Pi$ of policies, we define the \emph{optimal expected $T$-trial payoff with respect to $\Pi$} as
\[
G^*(T) \defeq \max_{\pi\in\Pi} G_\pi(T).
\]
For convenience, we also define the per-trial payoff of an algorithm or policy,
which is defined, respectively, by
\begin{eqnarray*}
g_\Aalg &\defeq& \frac{G_\Aalg(T)}{T} \\
g_\pi &\defeq& \frac{G_\pi(T)}{T} = \E_D\left[r_{1,\pi(\vecb{x}_1)}\right].
\end{eqnarray*}
Much research in multi-armed bandit problems is devoted to developing algorithms
with large total payoff.  Formally, we may search for an algorithm
minimizing \emph{regret} with respect to the optimal arm-selection
strategy in $\Pi$.  Here, the $T$-trial regret $R_\Aalg(T)$
of algorithm $\Aalg$ with respect to $\Pi$ is defined by
\begin{eqnarray}
R_\Aalg(T) \defeq G^*(T) - G_\Aalg(T). \label{eqn:regret-def}
\end{eqnarray}
An important special case of the general contextual bandit problem is the well-known \emph{$K$-armed bandit} in which the context $\vecb{x}_t$ remains constant for all $t$.  Since both the arm set and contexts are constant at every trial, they have no effect on a bandit algorithm, and so we will also refer to this type of bandit as a \emph{context-free} bandit.

In the example of news article recommendation, we may view articles in the pool as arms, and for the $t$-th user visit (trial $t$), one article (arm) is chosen to serve the user.  When the served article is clicked on, a payoff of $1$ is incurred; otherwise, the payoff is $0$.  With this definition of payoff, the expected payoff of an article is precisely its \emph{click-through rate (CTR)}, and choosing an article with maximum CTR is equivalent to maximizing the expected number of clicks from users, which in turn is the same as maximizing the total expected payoff in our bandit formulation.


\subsection{Existing Bandit Algorithms} \label{sec:existing-algorithms}

The fundamental challenge in bandit problems is the need for balancing
exploration and exploitation.  To minimize the regret in
\eqnref{eqn:regret-def}, an algorithm $\Aalg$ \emph{exploits} its
past experience to select the arm that appears best.  On the other
hand, this seemingly optimal arm may in fact be suboptimal, due to
imprecision in $\Aalg$'s knowledge.  In order to avoid this undesired
situation, $\Aalg$ has to \emph{explore} the world by actually choosing
seemingly suboptimal arms so as to gather more information about them (c.f.,
step~\ref{def:bandit-feedback} in the bandit process defined in the previous subsection).
Exploration can increase \emph{short-term} regret since some
suboptimal arms may be chosen.  However, obtaining information about
the arms' average payoffs (i.e., exploration) can refine $\Aalg$'s
estimate of the arms' payoffs and in turn reduce \emph{long-term} regret.
Clearly, neither a purely exploring nor a purely exploiting algorithm works
best in general, and a good tradeoff is needed.

There are roughly two classes of bandit algorithms.  The first class of
algorithms attempt to minimize the regret as the number of steps increases.
Formally, such algorithms $\Aalg$ ensure the quantity $R_\Aalg(T)/T$ vanishes
over time as $T$ grows.  While low-regret algorithms have been extensively
studied for the \emph{context-free} $K$-armed bandit problem~\cite{Berry85Bandit},
the more general contextual bandit problem has remained challenging.
Another class of algorithms are based on Bayes rule, such as Gittins index
methods~\cite{Gittens79}.
Such Bayesian approaches may have competitive performance with appropriate prior distributions,
but are often computationally prohibitive without coupling with approximation~\cite{Agarwal09Explore}.

The Appendix describes a few representative low-regret algorithms 
used in our experiments, but it should be noted that our method is
algorithm \emph{independent}, and so may be applied to evaluate
Bayesian algorithms as well.

\section{Unbiased Offline Evaluation} \label{sec:method}

Compared to machine learning in the more standard supervised learning setting,
evaluation of methods in a contextual bandit setting is frustratingly difficult.
Our goal here is to measure the performance of a {\em bandit algorithm} $\Aalg$,
that is, a rule for selecting an arm at each time step based on the
preceding interactions and current context (such as the algorithms described above).

More formally, we want to estimate the per-trial payoff
\[
g_\Aalg = \frac{G_\Aalg(T)}{T} = \frac{1}{T}\E_D\left[\sum_{t=1}^Tr_{t,a_t}\right] .
\]
Here, $a_t$ is the $t$-th action chosen by $\Aalg$, and in general depends on the previous contexts, actions, and observed rewards.  Because of the interactive nature of the problem, it would seem that
the only way to do this evaluation unbiasedly is to actually run the algorithm online on ``live''
data.
However, in practice, this approach is likely to be infeasible due to the
serious logistical challenges such as extensive engineering resources and potential risks on user experiences.
Rather, we may only
have {\em offline} data available that was collected at a previous
time using an entirely {\em different} logging policy.
Because payoffs are only observed for the arms chosen by the logging
policy, which are likely to differ from those chosen by the
algorithm $\Aalg$ being evaluated, it is not at all clear how to evaluate
$\Aalg$ based only on such logged data.
This evaluation problem may be viewed as a special case of the so-called
``off-policy policy evaluation problem'' in the reinforcement-learning
literature~\cite{Precup00Eligibility}.  In the multi-armed bandit setting, however, there is no need for ``temporal credit assignment'', and thus more efficient solutions are possible.

One solution is to build
a simulator to model the bandit process from the logged data, and then
evaluate $\Aalg$ with the simulator.  Although this approach is straightforward,
the modeling step is often very expensive and difficult,
and more importantly, it often 
introduces \emph{modeling bias} to the simulator, making it hard to justify
reliability of the obtained evaluation results.
In contrast, we propose an approach that is \emph{unbiased}, grounded
on data, and simple to implement.

In this section, we describe a sound technique for carrying out
such an evaluation, assuming that the individual events are
i.i.d., and that the logging policy
chose each arm at each time step uniformly at random.
Although we omit the details, this latter assumption can be
weakened considerably so that any randomized logging policy is allowed
and the algorithm can be modified accordingly using rejection
sampling, but at the cost of decreased data efficiency.
\lihong{Is it too obvious?  Or do we need to elaborate more on this extension?}

More precisely, we suppose that there is some unknown distribution $D$
from which tuples are drawn i.i.d.~of the form
$(\vecb{x}, 
r_{1}, \ldots, r_{K})$, each
consisting of \emph{observed} context and \emph{unobserved} payoffs for all arms.
We also posit access to a long sequence of logged events resulting
from the interaction of the uniformly random logging policy with the world.
Each such event consists of the context vector
$\vecb{x}$, 
a selected arm $a$, and the resulting observed payoff $r_a$.
Crucially, this logged data is partially labeled in the sense that only the payoff $r_a$ is observed for the single arm $a$
that was chosen uniformly at random.

Our goal is to use this data to evaluate a bandit algorithm
$\Aalg$, which is a (possibly randomized) mapping
for selecting the arm $a_t$ at time $t$ based on the history
$h_{t-1}$ of $t-1$ preceding events together with the current context.
Therefore, the data serves as a benchmark, with which people can evaluate and compare different bandit algorithms.  As in supervised learning, having such benchmark sets will allow easier, replicable comparisons of algorithms in real-life data.

It should be noted that this section focuses on contextual bandit
problems with constant arm sets of size $K$.  While this assumption
leads to easier exposition and analysis, it may not be satisfied in
practice.  For example, in the news article recommendation problem studied in
\secref{sec:application}, the set of arms is not fixed: new arms may
become available while old arms may be dismissed.  Consequently, the
events are independent but drawn from non-identical distributions.
We do not investigate this setting formally although it is possible to
generalize our setting in Section~\ref{sec:notation} to this variable arm set case. 
Empirically, we find the evaluator is very stable.

\subsection{An Unbiased Offline Evaluator}

In this subsection, for simplicity of exposition, we take this sequence
of logged events to be an \emph{infinitely} long stream.  But we also
give explicit bounds on the actual finite number of events required
by our evaluation method.  A variation for finite data streams is studied
in the next subsection.

The policy evaluator is shown in \algref{alg:pe}~\cite{Li10Contextual}.
The method takes as input a bandit algorithm $\Aalg$ and a desired number of
``valid'' events $T$ on which to base the evaluation.
We then step through the stream of logged events one by one.
If, given the current history $h_{t-1}$, it happens that the policy
$\Aalg$ chooses the same arm $a$ as the one that was selected by the
logging policy, then the event is retained (that is, added to the
history), and the total payoff $\hat{G}_\Aalg$ updated.
Otherwise, if the policy $\Aalg$ selects a different arm from the one
that was taken by the logging policy, then the event is entirely
ignored, and the algorithm proceeds to the next event without any
change in its state.

\begin{algorithm}[t]
\begin{algorithmic}[1]
\item Inputs: $T>0$; bandit algorithm $\Aalg$; stream of events $S$
\STATE $h_0\leftarrow\emptyset$ \COMMENT{An initially empty history}
\STATE $\hat{G}_\Aalg\leftarrow0$ \COMMENT{An initially zero total payoff}
\FOR{$t=1,2,3,\ldots,T$}
\REPEAT
\STATE Get next event $(\vecb{x},a,r_a)$ from $S$
\UNTIL{$\Aalg(h_{t-1},\vecb{x}) = a$}
\STATE $h_t \leftarrow \mbox{\sc concatenate}(h_{t-1}, (\vecb{x},a,r_{a}))$
\STATE $\hat{G}_\Aalg \leftarrow \hat{G}_\Aalg + r_{a}$
\ENDFOR
\STATE Output: $\hat{G}_\Aalg/T$
\end{algorithmic}
\caption{\algfont{Policy\_Evaluator} {\small(with infinite data stream).}} \label{alg:pe}
\end{algorithm}

Note that, because the logging policy chooses each arm uniformly at
random, each event is retained by this algorithm with probability
exactly $1/K$, independent of everything else.
This means that the events which are retained have the same
distribution as if they were selected by $D$.
As a result, we can prove that two processes are
equivalent: the first is evaluating the policy against $T$ real-world
events from $D$, and the second is evaluating the policy using the policy
evaluator on a stream of logged events.  \thmref{thm:simulation}
formalizes this intuition.

\begin{thm} \label{thm:simulation}
For all distributions $D$ of contexts and payoffs, 
all algorithms $\Aalg$, all $T$, all sequences of events $h_T$,
and all stream $S$ containing i.i.d. events from a uniformly random logging
policy and $D$, we have
\[
\Pr_{\mbox{Policy\_Evaluator}(\Aalg,S)}(h_T) = \Pr_{\Aalg,D} (h_T).
\]
Furthermore, let $L$ be the number of events obtained
from the stream to gather the length-$T$ history $h_T$, then
\begin{enumerate}
\item{the expected value of $L$ is $KT$, and}
\item{for any $\delta\in(0,1)$, with probability at least $1-\delta$,
$L \le 2K(T+\ln(1/\delta))$.}
\end{enumerate}
\end{thm}

This theorem says that {\em every} history $h_T$ has an identical probability
in the real world as in the policy evaluator.  Any statistics of these
histories, such as the estimated per-trial payoff $\hat{G}_\Aalg/T$ returned by \algref{alg:pe},
are therefore unbiased estimates of the respective quantities of the algorithm $\Aalg$.
Hence, by repeating \algref{alg:pe} multiple times and then averaging the returned
per-trial payoffs, we can accurately estimate
the total per-trial payoff $g_\Aalg$ of any algorithm $\Aalg$ and respective confidence intervals.
Further, the theorem guarantees that, with high probability, $O(KT)$ logged events are sufficient to retain a sample of size $T$.

\begin{proof}

The first statement can be proved by mathematical induction on the time steps of event streams~\cite{Li10Contextual}.  Second, since each event from the stream is retained with probability exactly
$1/K$, the expected number required to retain $T$ events is exactly $KT$.  Finally, the
high-probability bound is an application of the multiplicative form of Chernoff's inequality.
\qed
\end{proof}

Given the unbiasedness guarantee, one may expect concentration is also guaranteed; that is, the evaluator becomes more and more accurate as $T$ increases.  Unfortunately, such a conjecture is false for general bandit algorithms, as explained in Example~\ref{ex:no-convergence} of the next section.

\subsection{Sample Complexity Result}

Next, we consider a situation that may be more relevant to practical evaluation
of a \emph{static} policy when we have a \emph{finite} data set $S$ containing $L$
logged events.  Roughly speaking, the algorithm steps through every event in $D$
as in \algref{alg:pe} and obtains an estimate of the policy's average
per-trial payoff based on a \emph{random} number of valid events.
The detailed pseudocode in \algref{alg:pe2}.

\begin{algorithm}[t]
\begin{algorithmic}[1]
\item bandit algorithm $\Aalg$; stream of events $S$ of length $L$
\STATE $h_0\leftarrow\emptyset$ \COMMENT{An initially empty history}
\STATE $\hat{G}_\Aalg\leftarrow0$ \COMMENT{An initially zero total payoff}
\STATE $T\leftarrow0$ \COMMENT{An initially zero counter of valid events}
\FOR{$t=1,2,3,\ldots,L$}
\STATE Get the $t$-th event $(\vecb{x},a,r_a)$ from $S$
\IF{$\Aalg(h_{t-1},\vecb{x})=a$}
\STATE $h_t \leftarrow \mbox{\sc concatenate}(h_{t-1}, (\vecb{x},a,r_{a}))$
\STATE $\hat{G}_\Aalg \leftarrow \hat{G}_\Aalg + r_{a}$
\STATE $T \leftarrow T + 1$
\ELSE
\STATE $h_t \leftarrow h_{t-1}$
\ENDIF
\ENDFOR
\STATE Output: $\hat{G}_\Aalg/T$
\end{algorithmic}
\caption{\algfont{Policy\_Evaluator} {\small(with finite data stream)}.} \label{alg:pe2}
\end{algorithm}

\algref{alg:pe2} is very similar to \algref{alg:pe}.  The only difference is that the number of valid events, denoted $T$ in the pseudocode, is a random number with
mean $L/K$.  For this reason, the output of \algref{alg:pe2} (namely, $\hat{G}_\Aalg/T$)
may not be an unbiased estimate of the true per-trial payoff of $\Aalg$.
However, the next theorem shows that the final value
of $T$ will be arbitrarily close to $L/K$ with high probability as long as $L$
is large enough.  Using this fact, the theorem further shows that the returned value
of \algref{alg:pe2} is an accurate estimate of the true per-trial payoff with high probability when $\Aalg$
is a \emph{fixed} policy that chooses action $a_t$ independent of the history
$h_{t-1}$.  To emphasize that $\Aalg$ is a fixed policy, the following theorem and
its proof use $\pi$ instead of $\Aalg$.

\begin{thm} \label{thm:samplecomplexity}
For all distributions $D$ over contexts and payoffs, 
all policies $\pi$, all data stream $S$ containing $L$ i.i.d. events drawn
from a uniformly random logging policy and $D$, and all $\delta\in(0,1)$,
we have, with probability at least $1-\delta$, that
\[
\abs{\frac{\hat{G}_\pi}{T} - g_\pi} = O\left(\sqrt{\frac{Kg_\pi}{L}\ln\frac{1}{\delta}}\right).
\]
\end{thm}

Therefore, for any $g \ge g_\pi$, with high probability, the theorem
guarantees that the returned value $\hat{G}_\pi/T$ is a close estimate
of the true value $g_\pi$ with error on the order of
$\tilde{O}\left(\sqrt{Kg/L}\right)$.  As $L$ increases, the error
decreases to $0$ at the rate of $O(1/\sqrt{L})$.  This error bound
improves a previous result~\cite[Theorem~5]{Langford08Exploration} for a similar
offline evaluation algorithm and similarly provides a sharpened
analysis for the $T=1$ special case for policy evaluation in
reinforcement learning~\cite{Kearns00Approximate}.
\secref{sec:application} provides empirical evidence matching our
bound.

\begin{proof}
The proof involves a couple applications of the multiplicative
Chernoff/Hoeffding bound~\cite[Corollary~5.2]{McDiarmid89Method}.
To simplify notation, we use
$\Pr(\cdot)$ and $\E[\cdot]$ in the proof to denote the probability and
expectation with respect to randomness generated by $\pi$
and $S$.  Let $(\vecb{x}_{t},a_t,r_{t,a_t})$
be the $t$-th event in the stream $S$, $V_t$ be the (random)
indicator that $a_t$ matches the arm chosen by
policy $\pi$ in the context $(\vecb{x}_{t})$.
Then, $T=\sum_{t=1}^LV_t$, $\hat{G}_\pi=\sum_{t=1}^LV_tr_{t,a_t}$,
and the returned value of \algref{alg:pe2} is $\hat{G}_\pi/T$.
We bound the denominator and numerator, respectively.

First, since $a_t$ is chosen uniformly at random, we
have $\E[V_t]=1/K$ for all $t$ and thus $\E\left[\sum_{t=1}^LV_t\right]=L/K$.
Using the multiplicative form of Chernoff's bound, we have
\begin{eqnarray}
\Pr\left(\abs{T-\frac{L}{K}}
\ge\frac{\gamma_1L}{K}\right)\le2\exp\left(-\frac{L\gamma_1^2}{3K}\right)
\label{eqn:t-bound}
\end{eqnarray}
for any $\gamma_1>0$.  Let the right-hand side above be $\delta/2$
and solve for $\gamma_1$:
\[
\gamma_1 = \sqrt{\frac{3K}{L}\ln\frac{4}{\delta}}.
\]

Similarly, since $a_t$ is uniformly chosen, we have
$\E\left[\hat{G}_\pi\right] = Lg_\pi/K$.
Applying the multiplicative Chernoff bound again,
we have for any $\gamma_2>0$ that
\begin{eqnarray}
\Pr\left(\abs{\hat{G}_\pi-\frac{Lg_\pi}{K}}>\frac{\gamma_2Lg_\pi}{K}\right) \le 2\exp\left(-\frac{Lg_\pi\gamma_2^2}{3K}\right).
\label{eqn:g-bound}
\end{eqnarray}
Let the right-hand side above be $\delta/2$ and 
solve for $\gamma_2$:
\[
\gamma_2 = \sqrt{\frac{3K}{Lg_\pi}\ln\frac{4}{\delta}}.
\]

Now applying a union bound over the probabilistic statements in
\eqnsref{eqn:t-bound} and (\ref{eqn:g-bound}), we can see that,
with probability at least $1-\delta$, the following holds:
\begin{eqnarray*}
&& \qquad \frac{1-\gamma_1}{K} \le \frac{T}{L} \le \frac{1+\gamma_1}{K} \\
&& \frac{g_\pi(1-\gamma_2)}{K} \le \frac{\hat{G}_\pi}{L} \le \frac{g_\pi(1+\gamma_2)}{K}.
\end{eqnarray*}
These two inequalities together imply
\[
    \abs{\frac{\hat{G}_\pi}{T}-g_\pi}
\le \frac{(\gamma_1+\gamma_2)g_\pi}{1-\gamma_1}
= O\left(\sqrt{\frac{Kg_\pi}{L}\ln\frac{1}{\delta}}\right),
\]
which finishes the proof. \qed
\end{proof}

Given \thmref{thm:samplecomplexity}, one might wonder if a similar result holds
for general bandit \emph{algorithms}.  Unfortunately, the following example shows
that such a concentration result is impossible in general.

\begin{ex} \label{ex:no-convergence}
Consider a contextual bandit problem with $K=2$ and $x\in\{0,1\}$ in
which $r_{t,1}=1$ and $r_{t,2}=0$ for all $t=1,2,\ldots$.  Suppose $x$
is defined by a uniform random coin flip.  Let $\Aalg$ be an algorithm
that operates as follows: if $x_1 = 1$ the algorithm chooses $a_t=1$ for all
$t$; otherwise, it always chooses $a_t=2$.  Therefore, the expected
per-trial payoff of $\Aalg$ is $g_\Aalg=0.5$.  However, in any
\emph{individual} run of the algorithm, its $T$-step total reward
$\hat{G}_\Aalg$ is either $T$ (if $\Aalg$ always chooses $a_t=1$) or
$0$ (if $\Aalg$ always chooses $a_t=0$), and therefore,
$\abs{\hat{G}_\Aalg/T-g_\Aalg}\equiv0.5$ no matter how large $T$ is.
\end{ex}

This counterexample shows that an exponential tail style deviation
bound does not hold for general bandit algorithms that are
dependent on history.  Not all
hope is lost though---there are some known algorithms for which
deviation bounds are provable; for example, \algfont{epoch-greedy}
algorithm~\cite{Langford08Epoch}, \algfont{UCB1}~\cite{Auer02Finite},
and \algfont{EXP3.P}~\cite{Auer02Nonstochastic}.  Furthermore, as
commented earlier, we can always repeat the evaluation process
multiple times and then average the outcomes to get accurate estimate
of the algorithm's performance.  In the next section, we show
empirically that \algref{alg:pe} returns highly stable results for
all algorithms we have tried.

\section{Case Study} \label{sec:application}

\begin{figure}
\centering
\includegraphics[width=\columnwidth]{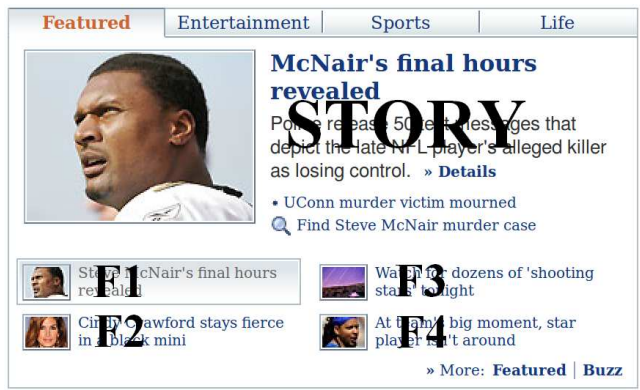}
\caption{A snapshot of the ``Featured'' tab in the Today Module on
the Yahoo!\ Front Page~\cite{Li10Contextual}. By default, the
article at F1 position is highlighted at the story position.}
\label{fig:fptoday}
\end{figure}

In this section, we apply the offline evaluation method in the
previous section to a large-scale, real-world problem with
variable arm sets to validate the effectiveness of our offline
evaluation methodology. Specifically, we provide empirical
evidence for: (i) the unbiasedness guarantee in
\thmref{thm:simulation}, (ii) the convergence rate in
\thmref{thm:samplecomplexity}, (iii) the low variance of the
evaluation result, and (iv) the effectiveness of the evaluation
method when the arm set may change over time.

While the proposed evaluation methodology has been applied to
the same application~\cite{Li10Contextual}, our focus here is on the
effectiveness of the offline evaluation method itself. More
importantly, we also provide empirical evidence of unbiasedness
for not only fixed policies but also learning algorithms, by
relating offline evaluation metric to online performance in
large-scale production buckets on Yahoo! Front Page.

We will first describe the application and
show how it can be modeled as a contextual bandit problem. Second,
we compare the offline evaluation result of a policy to its online
evaluation to show our evaluation approach is indeed unbiased and
it gives results that are asymptotically consistent when the
number of valid events (the quantity $T$ in
Algorithms~\ref{alg:pe} and \ref{alg:pe2}) is large. Third, we
provide empirical evidence that our offline evaluation method
gives very stable results for a few representative algorithms.
Finally, we study the relationship between offline evaluation results
to online bucket performance for three bandit algorithms.

\subsection{News Article Recommendation on Yahoo! Front Page Today Module}


The Today Module is the most prominent panel on the Yahoo!\ Front
Page, which is also one of the most visited pages on the Internet;
see a snapshot in Figure \ref{fig:fptoday}. The default
``Featured'' tab in the Today Module highlights four high-quality
news articles, selected from an hourly-refreshed article pool
maintained by human editors. As illustrated in Figure
\ref{fig:fptoday}, there are four articles at footer positions,
indexed by F1--F4. Each article is represented by a small picture
and a title. One of the four articles is highlighted at the story
position, which is featured by a large picture, a title and a
short summary along with related links. {By default, the article
at F1 is highlighted at the story position.} A user can click on
the highlighted article at the story position to read more details
if interested in the article. The event is recorded as a
story click. To draw visitors' attention, we would like to rank
available articles according to individual interests, and
highlight the most attractive article for each visitor at the
story position.  In this paper, we focus on selecting
articles for the story position.

This problem can be naturally modeled as a contextual bandit
problem.  Here, it is reasonable to assume each user visits
and their click probabilities on articles to be (approximately)
i.i.d.  Furthermore, each user has a set of features (such
as age, gender, etc.) from which the click probability of
a specific article may be inferred; these features are the contextual
information used in the bandit process.  Finally, we may view articles in the
pool as arms, and the payoff is $1$ if the user clicks on the
article and $0$ otherwise.  With this definition of payoff,
the expected payoff of an article is precisely its CTR,
, and choosing an article with
maximum CTR is equivalent to maximizing the expected number
of clicks from users, which in turn is the same as maximizing
the per-trial payoff $g_\pi$ in our bandit formulation.

We setup cookie-based buckets for evaluation. A bucket consists of a
certain amount of visitors. A cookie is a string of 13 letters
randomly generated by the web browser as an identifier.  We can
specify a cookie pattern to create a bucket. For example, we could let
users with the starting letter ``a'' in their cookies fall in one
bucket. In a cookie-based bucket, a user is served by the same policy,
unless the user changes the cookie and then belongs to another bucket.

For offline evaluation, millions of events were collected from a
``random bucket'' from Nov.~1, 2009 to Nov.~10, 2009. In the
random bucket, articles are randomly selected from the article
pool to serve users. There are about $40$ million
events in the offline evaluation data set, and about $20$ articles
available in the pool at every moment.

We focused on user interactions with the story article at the
story position only. The user interactions are recorded as two
types of events, user visit event and story click event. We chose
CTR as the metric of interest, which is defined as the ratio
between the number of story click events and the number
of user visits. 
To protect business-sensitive information,
we only report relative CTRs which are defined as the
ratio between true CTRs and a hidden constant.

\subsection{Unbiasedness Analysis} \label{sec:unbiased}

Given a policy, the unbiasedness of the offline evaluation
methodology can be empirically verified by comparing offline
metrics with online performance. We set up another cookie-based
bucket, noted as ``serving bucket'', to evaluate online
performance. In the serving bucket, a spatio-temporal algorithm~\cite{Agarwal09Spatio}
was deployed to estimate article CTRs.\footnote{Note that
the CTR estimates are updated every 5 minutes.}  The article with
the highest CTR estimate (also known as the winner article) was
then used to serve users. 
We extracted the serving policy from the
``serving bucket'', i.e., the best article at every 5 minutes from
Nov. 1 2009 to Nov. 10 2009. Note that it is in the same period of
time of the offline evaluation data set, ensuring that the
sets of available arms are the same in both the serving and random buckets.
Then, we used Algorithm \ref{alg:pe} to evaluate the serving policy on the
events from the random bucket for the offline metric.

It should be noted that the outcome of our experiments are not a
foregone conclusion of the mathematics presented, because the
setting differs in some ways from the i.i.d. assumption made in
our theorems as is typical in real-world applications.  In
particular, events are not exchangeable since old articles leave
the system and new ones enter, there are sometimes unlogged
business rule constraints on the serving policy, and users of
course do not behave independently when they repeatedly visit the
same site. We finesse away this last issue, but the first two are
still valid.

In the serving bucket, a winner article usually remains the best
for a while. During its winning time, the user repeatedly sees the
same article. At the same time, the users in the random bucket are
very likely to see different articles at user visit events, due to the
random serving policy. It is conceivable that the more a user views
the same article, the less likely the user clicks on the article.
This conditional effect violates the i.i.d. assumption in
Theorem~\ref{thm:simulation}. Fortunately, the discrepancy can be
removed by considering CTR on \emph{distinct views}. For each
user, consecutive events of viewing the same article are counted
as one user visit only. The CTR on distinct views in the serving
bucket measures user interactions to the winner articles across
the whole session. Regarding the offline evaluation metric as in
\algref{alg:pe2}, the subset of events sampled in the random
bucket also measures user interactions with the winner articles
across the whole session.

\begin{figure}[t]
\centering
\includegraphics[width=\columnwidth]{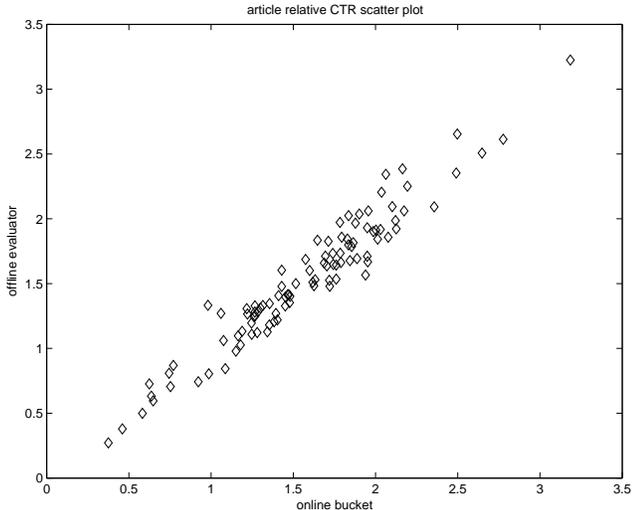}
\caption{Articles' CTRs in the online bucket versus offline estimates.}
\label{fig:articlectr}
\end{figure}

We first compared online and offline per-article CTRs.  Only winner
articles that were viewed more than $20,000$ times in the
serving bucket are used in the plot so that their online CTRs
are accurate enough to be treated as ground truth.
\figref{fig:articlectr} shows that the CTR
metric evaluated offline are very close to the CTR estimated
online.

\begin{figure}[h]
\centering
\includegraphics[width=\columnwidth]{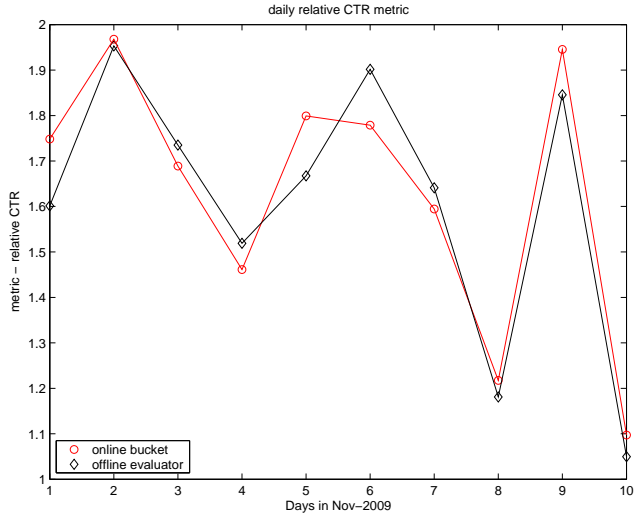}
\caption{Daily overall CTRs in the online bucket versus offline estimates.}
\label{fig:dailyctr}
\end{figure}

We next compared online and offline CTRs at the policy level.
These CTRs are the overall CTR of the serving policy aggregated
over all articles.  Figure \ref{fig:dailyctr} shows the two
CTRs are very close on each individual day.

Both sets of results corroborate the unbiasedness guarantee of
\thmref{thm:simulation}, a property of particular importance in practice
that is almost impossible with simulator-based evaluation methods.
Therefore, our evaluation method provides a solution that is accurate
(like bucket tests) without the cost and risk of running the
policy in the real system.

\subsection{Convergence Rate Analysis} \label{sec:errorbound}

We now study how the difference between offline and online CTRs
decreases with more data (namely, the quantity $T$ in the evaluation methods).
To show the convergence rate, we present the
estimated error versus the number of samples used in offline
evaluation.  Formally, we define the estimated error by
$e=\abs{c-\hat{c}}$, where $c$ and $\hat{c}$ are the true CTR
and estimated CTR, respectively.

\begin{figure}[t]
\centering
\includegraphics[width=\columnwidth]{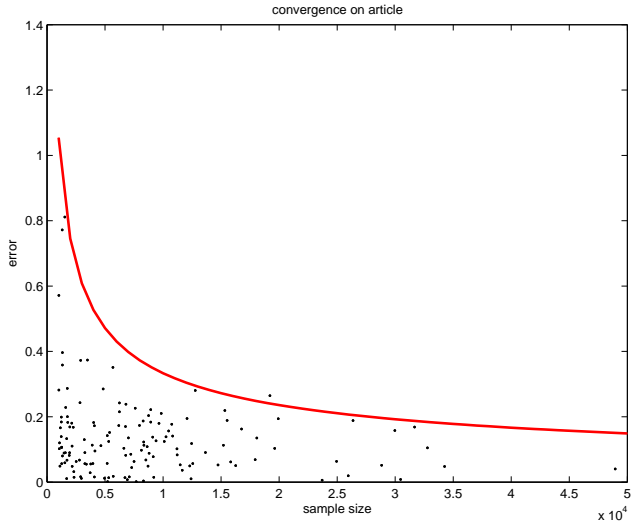}
\caption{Decay rate of error in articles' CTR estimates with increasing data size.
The red curve plots the function $1/\sqrt{x}$.}
\label{fig:articleconverge}
\end{figure}

\begin{figure}[t]
\centering
\includegraphics[width=\columnwidth]{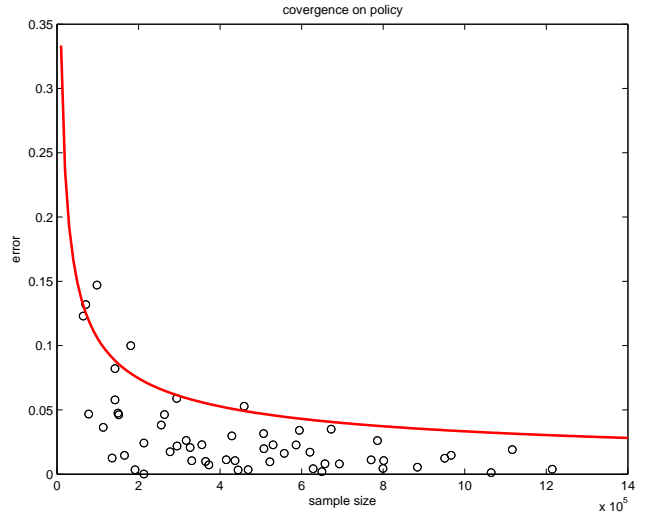}
\caption{Decay rate of error in overall CTR estimates with increasing data size.
The red curve plots the function $1/\sqrt{x}$.}
\label{fig:policyconverge}
\end{figure}

Figures~\ref{fig:articleconverge} and \ref{fig:policyconverge} present
convergence rate of the CTR estimate error for various articles and
the online serving policy, respectively, and the red curve is $1/\sqrt{T}$---the
functional form of the upper confidence bound.  These results suggest
that, in practice, we can observe the error decay rate predicted by
\thmref{thm:samplecomplexity} for reasonably stable algorithms such as
those evaluated.

\subsection{Low Variance of Evaluation Results}

In this subsection, we chose three representative algorithms (c.f., Appendix) to illustrate the low variance of the offline evaluation technique:
\begin{itemize}
\item{\algfont{$\epsilon$-greedy}, a stochastic, context-free algorithm;}
\item{\algfont{UCB}, a deterministic, context-free variant of \algfont{UCB1}~\cite{Auer02Finite};}
\item{\algfont{LinUCB}~\cite{Li10Contextual}, a deterministic, contextual bandit algorithm that uses ridge regression to estimate arm payoffs based on contexts.}
\end{itemize}

Each of the algorithms above has one parameter: $\epsilon$ for \algfont{$\epsilon$-greedy}
and $\alpha$ for \algfont{UCB} and \algfont{LinUCB} (see \cite{Li10Contextual} for
details).  We fixed the parameters to reasonable values: $\epsilon=0.4$ and $\alpha=1$.
We collected over $4,000,000$ user visits from a random bucket on May 1, 2009.
To evaluate variance, we subsampled this data so that each event is used
with probability $0.5$.  We ran each algorithm $100$ times on independently
subsampled events and measure the returned CTR using \algref{alg:pe2}.

Table~\ref{tab:stats} summarizes statistics of CTR estimates for the three
algorithms.\footnote{In the terminology of \cite{Li10Contextual}, the CTR
estimates reported in Table~\ref{tab:stats} are for the ``learning bucket''.
Similar standard deviations are found for the so-called ``deployment bucket''.}
It shows that the evaluation results are highly consistent across different
random runs.  Specifically, the ratio between standard deviation and the mean
CTR is about $2.4\%$ for \algfont{$\epsilon$-greedy}, and below $1.5\%$ for
the \algfont{UCB} and \algfont{LinUCB} which have known algorithm-specific deviation bounds.

This experiment demonstrates empirically that our evaluation method
can give results that have small variance for a few natural algorithms,
despite the artificial counterexample in \secref{sec:method},
suggesting that with large datasets the result obtained from only
one run of our evaluation method are already quite reliable.

\begin{table}[h]
\begin{center}
\begin{tabular}{l|cccc}
\hline\hline
algorithm & mean & std & max & min \\
\hline
\algfont{$\epsilon$-greedy} & $1.2664$ & $0.0308$ & $1.3079$ & $1.1671$ \\
\algfont{UCB} & $1.3278$ & $0.0192$ & $1.3661$ & $1.2812$ \\
\algfont{LinUCB} & $1.3867$ & $0.0157$ & $1.4268$ & $1.3491$ \\
\hline\hline
\end{tabular}
\end{center}
\caption{Statistics of CTR estimates for three representative algorithms using \algref{alg:pe2}.}
\label{tab:stats}
\end{table}

\vfill\eject

\subsection{Consistency with Online Performance}

\begin{figure}[t]
\centering
\includegraphics[width=\columnwidth]{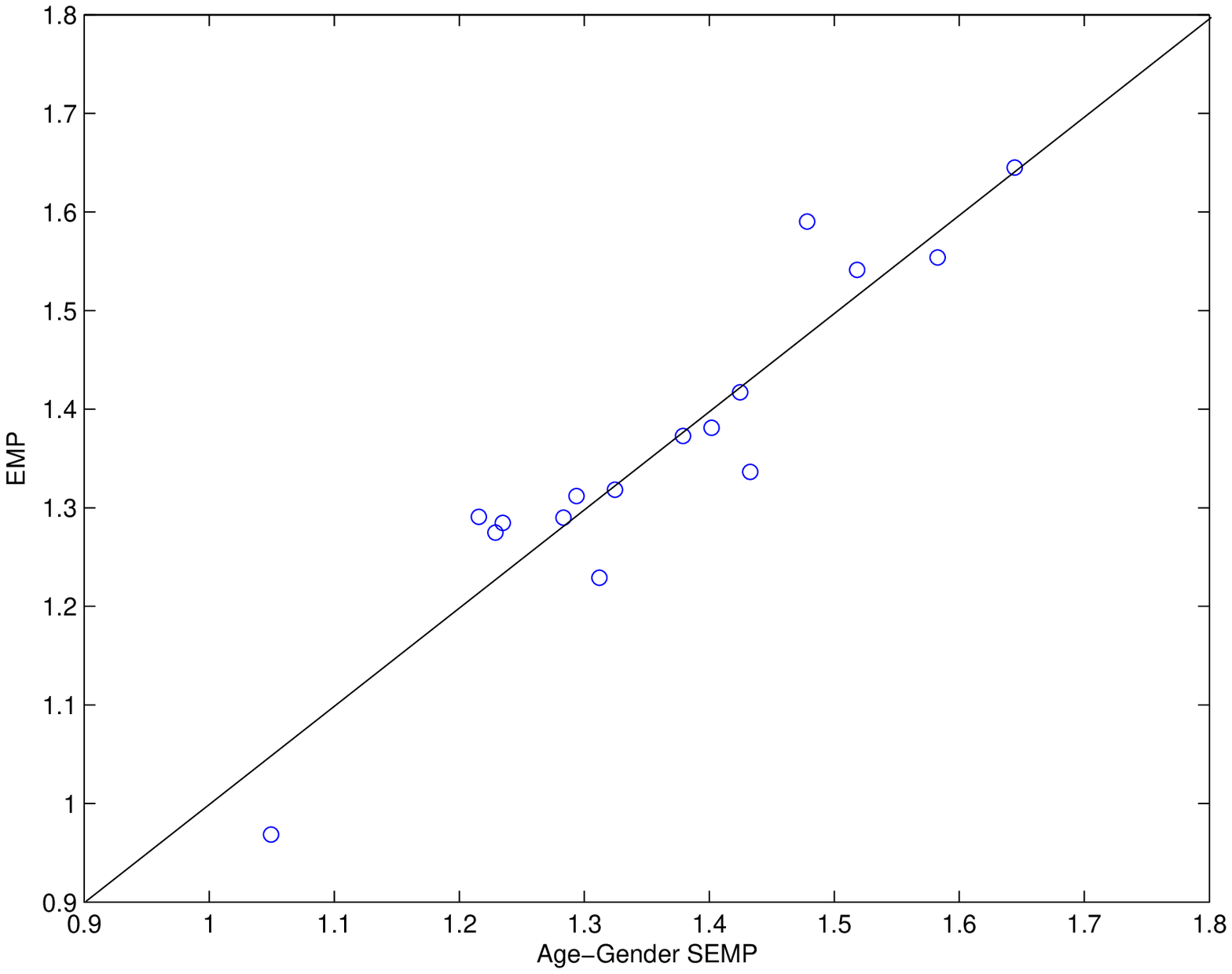}
\caption{Scatter plot of ratios of offline metric and online
bucket performance of 16 days in 2009.} \label{fig:semp}
\end{figure}

\begin{figure}[t]
\centering
\includegraphics[width=\columnwidth]{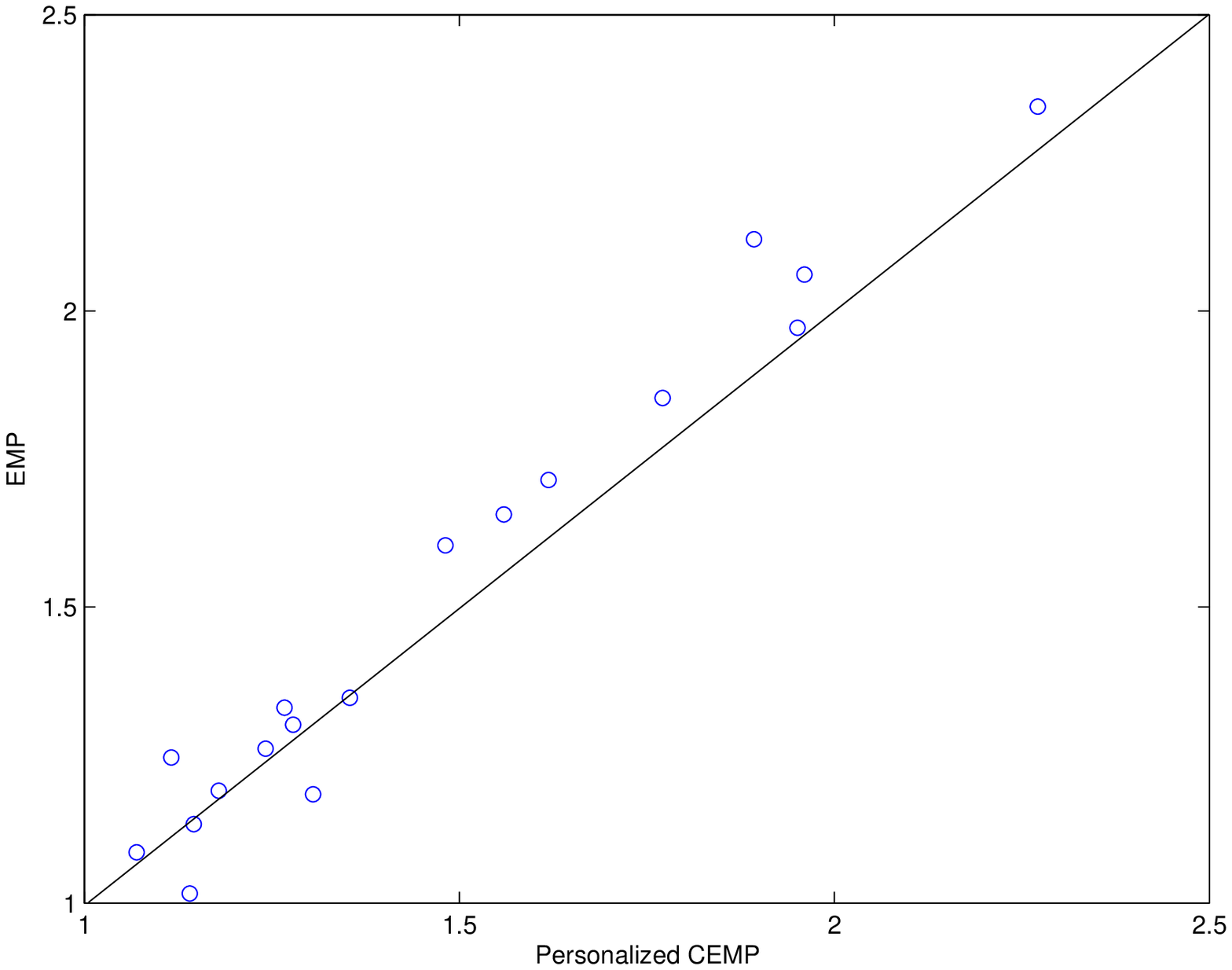}
\caption{Scatter plot of ratios of offline metric and online
bucket performance in 18 days in 2010.} \label{fig:cemp}
\end{figure}

Sections~\ref{sec:unbiased} and \ref{sec:errorbound} give evidence
for the accuracy when the offline evaluation method is applied to
a static policy that is fixed over time.  In this section, we show
complimentary accuracy results for \emph{learning algorithms} that
may be viewed as history-dependent, non-fixed policies.
In particular, we show the consistency between our offline evaluation
and online evaluation of three $\epsilon$-greedy bandit models:

\begin{itemize}

\item{Estimated Most Popular (\algfont{EMP}): we estimate CTR of available
articles over all users via a random exploration bucket, and then
serve users in the \algfont{EMP} bucket by the article of the highest CTR;}

\item{Segmented Most Popular (\algfont{SEMP}): we segment users into 18
clusters based on their age/gender information. We estimate CTR of
available articles within each cluster, and for each user cluster
serve the article with the highest CTR. Note that users'
feedback may change serving policy for the cluster in future
trials;}

\item{Contextual Bandit Model (\algfont{CEMP}): this is a fine-grained personalized model.
In this model, we define a separate context for each user based on
her age, gender, etc. For each available article, we maintain a
logistic regression model to predict its CTR given the user
context. When a user comes, in the \algfont{CEMP} bucket, we estimate the
CTRs of all articles for the user and select the article with
highest estimated CTR to display. Users with different contexts
may be served by different articles in this bucket, while each
user's feedback will affect other users' click probability
estimation on this article in future trials.}

\end{itemize}

For all these three bandit models, we set up three online
bcookie-based buckets to deploy the three bandit models
respectively. We also set up another bucket to collect random
exploration data. This random data is used to update the states of
three online bandit models and also used for our offline
evaluation. For a given period, we obtain the per-trial payoffs
$g_\Aalg^\mathrm{online}$ for $\Aalg\in\{\text{\algfont{EMP}, \algfont{SEMP}, \algfont{CEMP}}\}$.
Using data in the random exploration bucket, we run our offline
evaluation for these three models in the same period and get the
per-trial payoffs $g_\Aalg^\mathrm{offline}$.

It is important to note that there were unlogged business-rule
constraints in all online serving buckets of Today Module;
for instance, an article may be forced to shown in a given time window.
Fortunately, our data analysis (not reported here) suggested that such business rules
have roughly the same multiplicative impact on an algorithm's online
CTR, although this multiplicative factor may vary across different days.
To remove effects caused by business rules, we report the
ratio of offline CTR estimate and online CTR for each model:
$\rho_{\Aalg} = g_{\Aalg}^\mathrm{offline} / g_{\Aalg}^\mathrm{online}$.
If our offline evaluation metric is truthful of an algorithm's online metric
in the absence of business rules, then it is expected that,
for a given period of time like one day, $\rho_{\Aalg}$ should
remain constant ideally and does not depend on the algorithm $\Aalg$.

In Figure~\ref{fig:semp}, we present a scatter plot of $\rho_\text{\algfont{EMP}}$
vs. $\rho_\text{\algfont{SEMP}}$ of 16 days, from May 03, 2009 to May 18, 2009. In
each day, we have about $2,000,000$ views (i.e., user visits) in each of the four
online buckets. The scatter plot indicates a strong linear
correlation. The slope in least squares linear regression is $1.019$
and the standard deviation in residue vector is $0.0563$. We
observed that business rules give almost the same impact on CTR in
buckets for the two serving policies.

\algfont{SEMP} is a relatively simple bandit algorithm similar to \algfont{EMP}. In
the next experiment, we study the online/offline correlation of a
more complicated contextual bandit serving policy \algfont{CEMP}, in which
CTRs are estimated using logistic regression on user features and
a separate logistic regression model is maintained for each article.
Figure~\ref{fig:cemp} shows the scatter plot of
$\rho_\text{\algfont{EMP}}$ vs $\rho_\text{\algfont{CEMP}}$ in 18-day data from May 22, 2010 to June
8, 2010. In each day, we have $2,000,000\sim3,000,000$
views in each online bucket. The scatter plot again indicates a strong
linear correlation. In this comparison, the slope and standard
deviation in residue vector is $1.113$ and $0.075$ respectively.
It shows that the difference between our offline and online evaluation,
caused by business rules and other systemic factors, e.g. time-out in
user feature retrieval and delays in model update, is comparable across
bandit models.  Although the daily factor is unpredictable, the
relative performance of bandit models in offline evaluation is reserved in
online buckets.  Thus, our offline
evaluator can provide reliable comparison of different models on
historical data, even in the presence of business rules.

\section{Conclusions and Future Work} \label{sec:conclusions}

This paper studies an offline evaluation method of bandit algorithms
that relies on log data directly rather than on a simulator.
The only requirement of this method is that the log data is generated
i.i.d. with arms chosen by an (ideally uniformly) random policy.
We show that the evaluation method gives unbiased estimates of
quantities like total payoffs, and also provide a sample complexity
bound for the estimated error when the algorithm is a fixed policy.
The evaluation method is empirically validated using real-world data
collected from Yahoo! Front Page for the challenging application of online
news article recommendation.  Empirical results verify our theoretical
guarantees, and demonstrate both accuracy and stability of our method
using real online bucket results.  These encouraging results suggest
the usefulness of our evaluation method, which can be easily applied to
other related applications such as online refinement of ranking
results~\cite{Moon10Online} and ads display.

Our evaluation method, however, ignores $(K-1)/K$ fraction of
logged data. Therefore, it does not make use of \emph{all} data,
which can be a problem when $K$ is large or when data is expensive
to obtain.  Furthermore, in some risk-sensitive applications,
while we can inject some randomness during data collection, a
uniformly random policy might be too much to hope for due to
practical constraints (such as user satisfaction).  As we
mentioned earlier, our evaluation method may be extended to work
for data collected by any random policy with rejection sampling,
which enjoys similar unbiasedness guarantees, but reduces the data
efficiency at the same time.  An interesting future direction,
therefore, is exploiting problem-specific structures to avoid
exploration of the full arm space.  A related question is how
to make use of \emph{non-random} data for reliable offline
evaluation, for which a recent progress has been made~\cite{Strehl11Learning}.

\section*{Acknowledgements}

We appreciate valuable inputs from Robert Schapire.

\vfill\eject

\bibliographystyle{plain}
\bibliography{paper}


\section*{APPENDIX} 

One of the simplest and widely used algorithms is \algfont{$\epsilon$-greedy}~\cite{Sutton98Reinforcement}.
In each trial $t$, the algorithm first estimates the average payoff $\hat{\mu}_{t,a}$ of each arm $a$.  Then, with probability $1-\epsilon$, it chooses the \emph{greedy} arm that has the highest payoff estimate: $a_t=\arg\max_a\hat{\mu}_{t,a}$; with probability $\epsilon$, it chooses a random arm.  Clearly, each arm will be tried infinitely often in the limit, and so the payoff estimate $\hat{\mu}_{t,a}$ converges to the true value $\mu_a$ with probability $1$ as $t\rightarrow\infty$.  Furthermore, by decaying $\epsilon$ appropriately, the per-step regret, $R_\Aalg(T)/T$, converges to $0$ with probability $1$~\cite{Robbins52Some}.


The \algfont{$\epsilon$-greedy} strategy is \emph{unguided} since it picks a random arm for exploration.  Intuitive, when an arm is clearly suboptimal, it need not be explored.  In contrast, another class of algorithms generally known as ``upper confidence bound'' algorithms~\cite{Lai85Asymptotically,Auer02Finite} use a smarter way to balance exploration and exploitation.  In particular, in trial $t$, these algorithms estimate both the mean payoff $\hat{\mu}_{t,a}$ of each arm $a$ as well as a corresponding confidence interval $c_{t,a}$, so that $\abs{\hat{\mu}_{t,a}-\mu_a}\le c_{t,a}$ holds with high probability.  They then select the arm that achieves a highest upper confidence bound (UCB for short): $a_t = \arg\max_a\left(\hat{\mu}_{t,a}+\alpha c_{t,a}\right)$, where $\alpha$ is a tunable parameter that may increase slowly over time.  In other words, UCB algorithms choose an arm that either has a high payoff estimate, or a high estimation uncertainty measure (corresponding to large values of $\alpha c_{t,a}$).  As more data have been collected to refine the payoff estimate, the confidence interval vanishes, and the algorithms will behave more greedily.  With appropriately defined confidence intervals and parameter $\alpha$, it can be shown that such algorithms have a small total $T$-trial regret that is only logarithmic in the total number of trials $T$~\cite{Lai85Asymptotically,Auer02Finite}.

While context-free $K$-armed bandits are extensively studied and well
understood, the more general contextual bandit problem has largely remained
open.  The \algfont{EXP4} algorithm and its variants~\cite{Auer02Nonstochastic,Beygelzimer11Contextual}
use the exponential weighting technique to achieve an
$\tilde{O}(\sqrt{T})$ regret in expectation, where
$\tilde{O}(x)\defeq O(x \ln x)$, even if the sequence of contexts and payoffs
are chosen by an adversarial world, but the computational complexity
may be exponential in the number of features in general.  Another
general contextual bandit algorithm is the \algfont{epoch-greedy}
algorithm~\cite{Langford08Epoch} that is similar to
\algfont{$\epsilon$-greedy} with adaptively shrinking $\epsilon$.
Assuming the sequence of contexts, $\vecb{x}_1,\ldots,\vecb{x}_T$, is
i.i.d., this algorithm is computationally efficient given an oracle
empirical risk minimizer but has the weaker regret guarantee of
$\tilde{O}(T^{2/3})$ in general, with stronger guarantees in various
special cases.

Algorithms with stronger regret guarantees may be designed under various modeling assumptions about the contextual bandit.  Assuming the expected payoff of an arm is linear in its features (namely, $\E_D[r_{t,a}\mid\vecb{x}_{t,a}]=\vecb{w}^\mt \vecb{\vecb{x}_{t,a}}$ for some coefficient vector $\vecb{w}$), both \algfont{LinRel}~\cite{Auer02Using} and
\algfont{LinUCB}~\cite{Li10Contextual,Chu11Contextual,Dorard09Gaussian} are essentially UCB-type approaches generalized
to linear payoff functions, and their variants have a regret of $\tilde{O}(\sqrt{T})$,
a significant improvement over earlier algorithms~\cite{Abe03Reinforcement} as well as the more general \algfont{epoch-greedy} algorithm.  Extensions to generalized linear models~\cite{Filippi11Parametric} are also possible and can still enjoy the same $\tilde{O}(\sqrt{T})$ regret guarantee.


\end{document}